\documentclass[letterpaper, 10 pt, conference]{ieeeconf}

\IEEEoverridecommandlockouts                              

\usepackage{times}

\usepackage{algorithmic}
\usepackage{algorithm}
\usepackage{latexsym,amssymb,amsmath,amsfonts}
\usepackage[english]{babel}
\usepackage{bm}
\usepackage{graphicx}
\usepackage{subcaption}

\usepackage[usenames]{color}
\usepackage{tikz}

\usetikzlibrary{shapes.geometric, arrows,backgrounds}
\usetikzlibrary{positioning,fit,calc}
\usepackage{scalefnt}

 \title{\LARGE \bf
 	Can Co-robots Learn to Teach?
 }
 
 \author{Harshal Maske, Emily Kieson, Girish Chowdhary, and Charles Abramson
 	\thanks{Harshal Maske, and Asst. Prof. Girish Chowdhary$^*$ are with the Distributed Autonomous Systems laboratory (DASLAB), University of Illinois at Urbana Champaign,{\tt \{hmaske2@illinois.edu, girishc@illinois.edu\}}. Emily Kieson, and Prof. Charles Abramson are with the Comparative Psychology Laboratory, Oklahoma State University, {\tt \{kieson@okstate.edu,  charles.abramson@okstate.edu\}}. }
 }
 
  \graphicspath{{figures/}}
  
\newcommand{\A}{\mathcal{A}}
\newcommand{\actelem}{r}
\newcommand{\Njoints}{n}

\newcommand{\tindex}{\tau}
\newcommand{\Time}{T}

\newcommand{\actprim}{a}
\newcommand{\assignInd}{z}
\newcommand{\subgStateAct}{\mathcal{S}}
\newcommand{\obsset}{\mathbf{\textit{O}}}
\newcommand{\posvec}{{x}}
\newcommand{\vvec}{{v}}
\newcommand{\evec}{s^e}
  
\begin{document}
  	
  	\maketitle
  	\thispagestyle{empty}
  	\pagestyle{empty}
  	
  	\begin{abstract}
  		We explore beyond existing work on learning from demonstration by asking the question: ``Can robots learn to teach?", that is, can a robot autonomously learn an instructional policy from expert demonstration and use it to instruct or collaborate with humans in executing complex tasks in uncertain environments?
  		In this paper we pursue a solution to this problem by leveraging the idea that humans often implicitly decompose a higher level task into several subgoals whose execution brings the task closer to completion. We propose Dirichlet process  
  		based non-parametric Inverse
  		Reinforcement Learning (DPMIRL) approach for reward based unsupervised clustering of task space into subgoals. This approach is shown to capture the latent
  		subgoals that a human teacher would have utilized to train a novice. The notion of ``action primitive'' is introduced as the means to communicate instruction policy to humans in the least complicated manner, and as a computationally efficient tool to segment demonstration data. We evaluate our approach through experiments on
  		hydraulic actuated scaled model of an excavator and evaluate
  		and compare different teaching strategies utilized by the robot.

  	\end{abstract}

\section{Introduction}

In many real world robotic applications, human operators play a critical role in ensuring the safety and efficiency of the task. Some examples include heavy construction and agricultural robotics where human operators of co-robots such as excavators, tractors, and backhoes must make safety-critical decisions in real-time under uncertain and dynamically changing environments. The \textit{skill-gap} between expert and novice operators of these robots is a significant limiting factor in ensuring safety, efficiency, and quality at work-sites.  If a co-robot was able to learn from experts and utilize that knowledge to assist or teach novice operators, significant performance gains could be achieved. In this paper, we study the crucial problem of directly learning instruction policies for novice operators from demonstrations provided by skilled operators. 
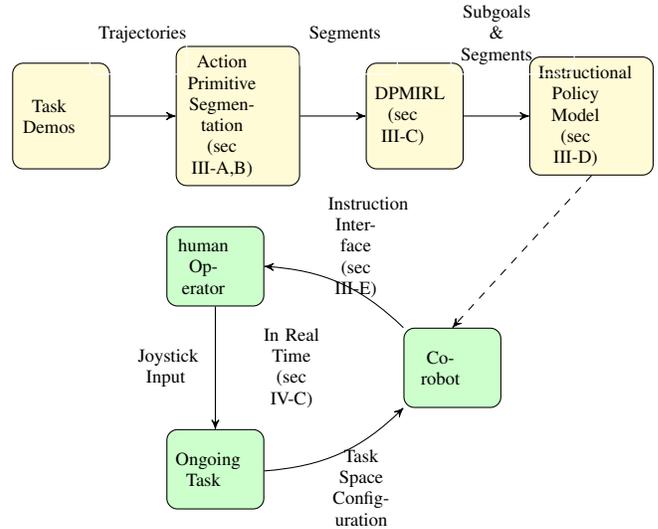
\begin{figure}[tbh]
	\centering
	\begin{tikzpicture}[node distance=18.5mm, auto,font=\scriptsize,->,>=stealth']
	
	
	\tikzstyle{blockb} = [rectangle, draw, fill=yellow!20, 
	text width=4em, text centered, rounded corners, minimum height=4em]
	\tikzstyle{blockbg} = [rectangle, draw, fill=green!20, 
	text width=4em, text centered, rounded corners, minimum height=4em]
	\tikzstyle{blocks} = [rectangle, draw, fill=yellow!20, 
	text width=3em, text centered, rounded corners, minimum height=4em]
	\tikzstyle{blocksg} = [rectangle, draw, fill=green!20, 
	text width=3em, text centered, rounded corners, minimum height=3em]
	
	\tikzstyle{blockw} = [rectangle, draw=white, 
	text width=3em, text centered, rounded corners, minimum height=3em]
	\tikzstyle{blockww} = [rectangle, draw=white, 
	text width=4em, text centered, rounded corners, minimum height=3em]
	
	\node [blocks] (demo) {Task Demos};
	
	\node [blockb, right of=demo, xshift=5mm] (segment) {Action Primitive Segmentation (sec \ref{sec:primitive},B)};
	
	\node [blocks, right of=segment, xshift=5mm] (dpmirl) {DPMIRL (sec \ref{sec:dpmirl})};
	
	\node [blockb, right of=dpmirl, xshift=5mm] (policy) {Instructional Policy Model (sec \ref{sec:instructionmodel})};
	
	\node[above=of demo,yshift=-18mm] (dummy2) {};
	\node[blockw, right=of dummy2,xshift=-16mm,yshift=2mm] (traj) {Trajectories};
	
	\node[blockw, right=of dummy2,xshift=12mm,yshift=2mm] (seg) {Segments};
	
	\node[blockww, right=of dummy2,xshift=32mm,yshift=2mm] (seg) {Subgoals \& Segments};
	
	\path[every node/.style={font=\sffamily\small}]
	(demo) edge node [right] {} (segment)
	(segment) edge node [right] {} (dpmirl)
	(dpmirl) edge node [right] {} (policy);
	
	\node [blocksg, below of=policy,anchor=west,xshift=-25mm,yshift=-15mm] (teach) {Co-robot};
	\node[above=of teach] (dummy) {};
	
	\node [blocksg, below of=teach,anchor=east,xshift=-25mm,yshift=5mm] (task) {Ongoing Task};
	\node [blocksg, above of=teach,anchor=east,xshift=-25mm,yshift=-5mm] (human) {human Operator};
	
	\node[blockw, left=of dummy,xshift=15mm,yshift=-9mm] (interf) {Instruction Interface (sec \ref{sec:interface})};
	\node[blockww, left=of dummy,xshift=18mm,yshift=-41mm] (config) {Task Space Configuration};
	\node[blockw, left=of teach,xshift=-5mm,yshift=0mm] (config) {Joystick Input};
	\node[blockww, left=of teach,xshift=14mm,yshift=0mm] (tim) {In Real Time (sec \ref{sec:testing})};
	
	\path (task) edge[->, bend right=20] (teach.south west);
	\path (teach.north west) edge[->, bend right=20] (human.east);
	\path (task.north) edge[<-, bend right=0] (human.south);
	
	\draw [dashed,->] (policy.south)--(teach.north);
	\end{tikzpicture}
	\caption{Co-robot learns instructional policy from expert demonstrations (off-line learning in yellow). In real time (shown in green), co-robot generates instruction/guidance for human operators based on the current task space configuration and the  instructional policy model. }
	\label{fig:flowchart}
\end{figure}
Learning from Demonstration has been widely studied in the context of robots learning to do a task from teacher demonstrations \cite{argall2009lfd}. However, when a robot needs to teach a human operator, the robot needs to do much more than just learning to imitate the demonstrated task. Rather, it has to simplify and decompose the tasks into human understandable task-primitives, and communicate back the essential sequence of actions to guide the human learning from the robot. This brings us to the very crucial question: How can Robots learn to Teach? We argue that there are two important aspects to answering this question: First is the development of practical algorithms that would allow a co-robot to extract latent subgoals or skills that a human teacher would have utilized to instruct other humans. Second, is the development of feedback strategies for providing the appropriate task specific guidance to the human based on the current task state. The  approach formulated in this paper is designed to address both these aspects and is shown through extensive experimentations to enable robots to teach complex tasks to human operators (see figure \ref{fig:flowchart}). 


The main contribution of this paper is a method to directly learn an instructional policy from human demonstrations. We define an instructional policy as a feedback policy that utilizes the robots current and past state to suggest the best set of future actions in order to proceed with a given task. This should be contrasted with existing LfD work that has focused primarily on the robot learning a policy for executing the task by itself. 
Yet, our approach is highly scalable and generalizable, and has been demonstrated to work on a realistic LfD problem with multiple degrees of freedom and uncertain operating conditions. Hence, it can also be used in a pure LfD form for complex real-world robotic tasks, such as those often encountered in construction. To ensure scalability and generalizability as well as to simplify communication with the human learner, we introduce the notion of \textit{action primitives} which are used in unsupervised segmentation of demonstration trajectories.  Unlike the motion primitives, action primitives are defined in the joint space of a robot, and are hence universally generalizable to any articulated robot, allowing explicit segmentation of demonstration data without the need for computationally expensive and time consuming MCMC sampling that has been utilized by existing LfD approaches. 


The developed approach is demonstrated in a rigorously designed large human-robot experiment with $ 113 $ human participants learning to perform the truck loading task with a scaled excavator in a different configuration (each time). Demonstrations from experts were collected and utilized by our algorithms to learn the instructional policy. Two different interfaces of communicating the correct action primitives to the human operators are investigated, and shown to lead to statistically significant improvement in the human learner's task completion time, task safety, and retention. Although all of the experiments are done with a scaled excavator, the results should translate well to working with real excavators, and have immediate significance for teleoperated robotics. In fact,  we note that working with a scaled excavator can be more cognitively demanding than working from within a real excavator because the operator has to transpose themselves to the correct reference frame as the turret rotates. 


 \section{Background}
 \subsection{Learning from Demonstration}
 Much of the existing work in LfD has focused on the problem of enabling the robot to learn a closed loop policy to perform a set of actions on its own. 
 LfD has been successful in teaching robots tennis swings \cite{schaal2006dynamic}, walking gaits \cite{nakanishi2004learning}, and complex helicopter maneuvers \cite{abbeel2004apprenticeship}. A review of various LfD techniques is available in \cite{argall2009lfd}. 
 In a subset of methods called Inverse Reinforcement Learning (IRL), the reward function is learned from a set of expert demonstrations \cite{abbeel2004apprenticeship}. Learning reward is argued by \cite{michini2015bayesian} to be equivalent to high-level description of the task, that explains the expert's
 behavior in a richer sense than the policy alone.

 Some recent work in LfD has focused on performing automatic segmentation of demonstrations into simpler reusable primitives \cite{konidaris2011robot, grollman2010incremental, butterfield2010learning, niekum2012learning}. The work by \cite{niekum2013incremental} uses the Beta Process Autoregressive Hidden Markov Model (BP-AR-HMM) developed by \cite{fox2011joint} to perform  auto-segmentation of time series data available from multiple demonstrations. For sequencing \cite{butterfield2010learning, grollman2010incremental} developed novel non-parametric probabilistic models in contrast \cite{niekum2013incremental} developed a finite state automaton that utilizes pose information of task objects and segment lengths for sequencing. In these methods, the segments or task primitives are loosely defined, with no bounds on the number of possible segments, thus limiting their re-usability. In this paper, we first propose a definition for segments, that is then utilized to perform segmentation of an unstructured demonstration. Moreover, this procedure does not rely on computationally inefficient Gibbs sampling for learning model parameters.  
 Aforementioned algorithms attempt to directly learn the policy; next we discuss reward based IRL techniques for LfD. Our attempt is to leverage key concepts from these two school of thoughts to develop algorithm for inverse LfD to enable robots to assist or instruct humans. 
 
 \subsection{Bayesian Nonparametric Inverse Reinforcement Learning}\label{sec:bnirl}
 Inverse Reinforcement Learning (IRL) is an LfD technique concerned with finding hidden reward function of an expert human demonstrator from
 the demonstrated state and action samples \cite{ziebart2008maximum}, \cite{abbeel2004apprenticeship}. Recently, an approach to solve IRL by automatically
 decomposing the reward function into a series of sugoals, which was viewed as local reward functions,
 was proposed \cite{michini2015bayesian}. We utilize the notion of executing subgoals in a particular sequence to perform a task as a key component of instruction policy model. This is based on research that deals with human expertise in complex environment \cite{nokes:2010:problem}, \cite{harre2011development}. According to these findings, humans often form implicit decompositions of higher level tasks into several subgoals, so that the execution of each subgoal brings the task closer to completion. 
 
 In IRL \cite{abbeel2004apprenticeship} a Markov Decision Process (MDP) without the reward function $ R(s) $ i.e. $ MDP\backslash R $ is given. A demonstration set $ \obsset $ consists of state action pairs, $ \obsset = \{(s_1,a_1),\dots,(s_N,a_N)\}$, where each pair $ \obsset_i = (s_i,a_i) $ indicates that the action $ a_i $ was performed from the state $ s_i $. Multiple set of demonstrations are used as an input to the IRL algorithm to obtain an estimate of reward function $ \hat{R}(s) $, such that the corresponding optimal policy $ \pi^* $ matches the observations. The IRL problem is ill-posed, since defining reward as, $ \hat{R}(s) = c ~ \forall s \in S $, would make any set of state-action pairs trivially optimal. Moreover, it is possible to encounter dissimilar actions from a particular state $ s_i $. This ambiguity was resolved by restricting the reward function to be of certain form \cite{ratliff2006maximum, syed2007game, neu2012apprenticeship}. Later \cite{ramachandran2007bayesian} developed a standard Bayesian inference procedure to learn reward function. 
 
 The IRL methods cited above attempt to explain the
 entire observations set $ \obsset $ with a single, complex reward
 function  resulting in a large computational burden when the space of candidate reward functions is large. To overcome this limitation, Michini et. al. \cite{michini2015bayesian} developed Bayesian non-parametric IRL (BNIRL) that partitions the demonstration set $ \obsset $ and explains each partition with a simple reward function or a ``subgoal'' $ R_g(s) $, which consists of a positive reward at a single coordinate $ g $ in the state (or feature) space (zero elsewhere). A Dirichlet process prior (Chinese Restaurant process (CRP) construction) is assumed over the unknown number of partitions that consists of observed state-action pairs $ \obsset_i \equiv (s_i,a_i) $. Partition assignment for each observation $ \obsset_i $ is denoted by $ z_i $. Posterior over the partition assignment is given by 
 \begin{equation}\label{eq:posterior}
 P(z_i|z_{-i}, \obsset) \propto \underbrace{P(z_i|z_{-i})}_{CRP} ~ \underbrace{P(\obsset_i|R_{z_i})}_{likelihood}  
 \end{equation} 
 where the first term denotes standard CRP prior and the second term evaluates the likelihood of the action $ a_i $ given the subgoal reward function $ R_{z_i} $ corresponding to partition (or subgoal) identified by $ z_i $. This likelihood term is evaluated using exponential rationality model (similar to that in \cite{ramachandran2007bayesian}):
 \begin{equation}\label{eq:act_lkl}
 P(\obsset_i|R_{z_i}) = P(a_i|s_i,z_i) \propto e^{\alpha Q^*(s_i,a_i,R_{z_i})}
 \end{equation}
 where the parameter $ \alpha $ represents the degree of confidence in the demonstrator's ability to maximize reward. The evaluation of optimal action value function $ Q^* $ requires substantial computation and becomes infeasible for large state spaces. Hence, the author developed an approximation based on action comparison for the action likelihood (\ref{eq:act_lkl}), as follows
 
 \begin{equation}\label{eq:act_comp}
 P(\obsset_i|R_{z_i}) = P(a_i|s_i,z_i) \propto e^{\alpha ||a_i - a_{CL}||_2}
 \end{equation}  
 
 where $ a_{CL} $ is the action given by some closed-loop controller attempting to go from state $ s_i $ to subgoal $ g_{z_i} $. Note that $ z_i $ is a partition assignment variable for state $ s_i $, if $ z_i = k $, then $ g_k $ is the coordinate of $ k^{th} $ subgoal. This approximation to BNIRL \cite{michini2013scalable}, enables successful application of IRL to real-world learning scenario characterized by large state space such as quad-rotor domain. However, their approach was tested for a very basic maneuver of traversing four way-points in space. We utilize the structure of BNIRL, particularly the partitioning of high dimensional state space into finite subgoals, and propose a simplification that extends its application to more complex tasks in real world settings. 

 \subsection{Dirichlet Process Means for Gaussian mixture model}\label{sec:dpmeans}
 Dirichlet process (DP) is a well known prior for the parameters of a Gaussian mixture model when the number of mixture components are not known a-priori. Recently Kulis and Jordan \cite{kulis2012revisiting} have shown that the Gibbs sampling algorithm for the Dirichlet process
 mixture approaches a hard clustering algorithm when the covariances of the Gaussian variables tend to zero. This results in a k-means-like clustering objective given by
 \begin{eqnarray}
 & \min \sum_{c=1}^{k} \sum_{x\in l_c}||x-\mu_c||^2 + \lambda k \nonumber \\
 &\text{where} \quad \mu_c = \frac{1}{|l_c|} \sum_{x\in l_c} x \nonumber
 \end{eqnarray} 
 where $ k $ is the number of clusters, $ \mu_c $ is the mean for cluster $ c $, and $ l_c $ is the set of data points $ x $ that belongs to the cluster $ c $.
 This objective function is similar to that of k-means but includes a penalty $ \lambda $ for the number of clusters. DP-means algorithm behaves similarly to that of k-means with the exception
 that a new cluster is formed whenever a point is
 farther than $ \lambda $ away from every existing cluster centroid. The algorithm is initialized with a single cluster whose mean is simply the global centroid. We use this approach to cluster the states by assuming Gaussian distribution over their euclidean norm. 
 
 \section{Methodology}
 In LfD literature dynamic movement primitive (DMP) framework \cite{ijspeert2002movement} is a prevalent approach to imitate demonstrations. Given the start and goal position in task space, a robot uses the DMP to generate trajectory (or set of states $ s $) for end-effector position and its mechanism, which is then mapped to appropriate joint angle (or set of actions $ a $) using inverse kinematics \cite{pastor2009learning} and thus the entire state-action policy map.  We are  interested to generate instruction policy for operators who perform the task, for which DMPs, which are encoded in terms of the trajectory, may not be the best solution. To track a given trajectory (i.e. the set of states $ s $) in 3-D space is a complex task for an operator and therefore can present difficulties in teaching. On the other hand, instructions in terms of joystick movements are easily understandable by the operator and are arguably the simplest means of communicating desired actions to operators. So why can't we use the joint angles i.e. the set of actions $ a $ from DMP? We could, since the joint angles can be mapped to joystick movement, however in order to do so, we need to perform two complex transformations ($ s $ to $ a $ and $ a $ to joystick actuations). Furthermore, it is not desirable to continuously instruct joystick movement to reach the goal from a start position. To tackle this issue, we introduce the notion of action primitives designed to simplify robot to human teaching tasks.
 
 \subsection{Action Primitives}\label{sec:primitive}
 The definition of action primitives is developed for articulated robots, however, it should be generalizable to other types of robots. For an articulated robot, a revolute joint provides single-axis rotation function, hence a joint $ j $ can take three broadly defined possible states: i) counterclockwise rotation, ii) stationary or non-zero noisy perturbation, or iii) clockwise rotation. Note that it is possible to add further resolution to this decomposition.  Let $ \actelem_j $ be the variable that takes on values from the set $ \{1,2,3\} $ corresponding to these states respectively. Thus $ \actelem_j=1 $ implies that the joint $ j $ is rotating counterclockwise and so on. The assignment of values $ \{1,2,3\} $ is arbitrary. \textit{Mathematically, an action primitive defines the action of a robot in terms of variable $ \actelem_j $ for each joint $ j $}. Action primitives are defined in the joint space of a robot and hence map directly to joystick movements. Secondly, in this formulation, an action primitive changes only when the direction of rotation, i.e. the variable $ \actelem_j $ changes for a joint $ j $. This has two significant consequences: First an action primitive can be used to decompose a demonstration consisting of continuous state action spaces into finitely many discrete state-action pairs, this is the basis for segmentation approach discussed in next section. Second, known  sequence of these finite number of action primitives can be used to generate step-wise instruction policy model to guide humans (discussed further in section \ref{sec:instructionmodel}).
 An example of action primitive $ \actprim_\tindex $ at a time $ \tindex $ for a robot with $ \Njoints = 3 $ revolute joints is $  \actprim_\tindex = [\actelem_1 = 1; \actelem_2 = 2; \actelem_3 = 2]$ or just $ \actprim_\tindex = [1, 2, 2]^T $ indicative of counterclockwise rotation for the first revolute joint. Clearly for a robot with $ \Njoints $ joints, there would exist finitely many action primitives, as opposed to infinitely many joint velocities obtained from DMPs.
 \subsection{Action Primitive based Segmentation of Task demos}\label{sec:segmentation}
 Action primitives are closely related to the joint space of a robot and is a least complicated mechanism to communicate instructions to humans when compared to dynamic motion primitives or any other existing methods because they relate directly to joystick motion. We now describe how action primitives can be used for segmenting demonstrations. Demonstration data in the form of joint positions $  \posvec_\tindex  $, and joint velocities $ \vvec_\tindex $ sampled from a continuous demonstration of a task at time instants $ \tindex = 1,2,\dots,\Time $, is used as an input to the segmentation algorithm.  Classification of sampled velocities into action primitives generates the required segmentation. Each segment will be an action primitive that spans over finite instants of time. But first we need to define distribution for each action primitive class. Recall that an action primitive is defined by the value of its variable $ \actelem_j $ for each joint $ j $. As noted earlier each $ \actelem_j \in \{1,2,3\} $, hence we need three distributions for each joint $ j $ of the robot. To accomplish this, we cluster sampled velocities assuming Gaussian distribution for each cluster. 
 
\begin{figure}[tbh]
 	\centering
 	\includegraphics[width=3.5in,height=1.8in]{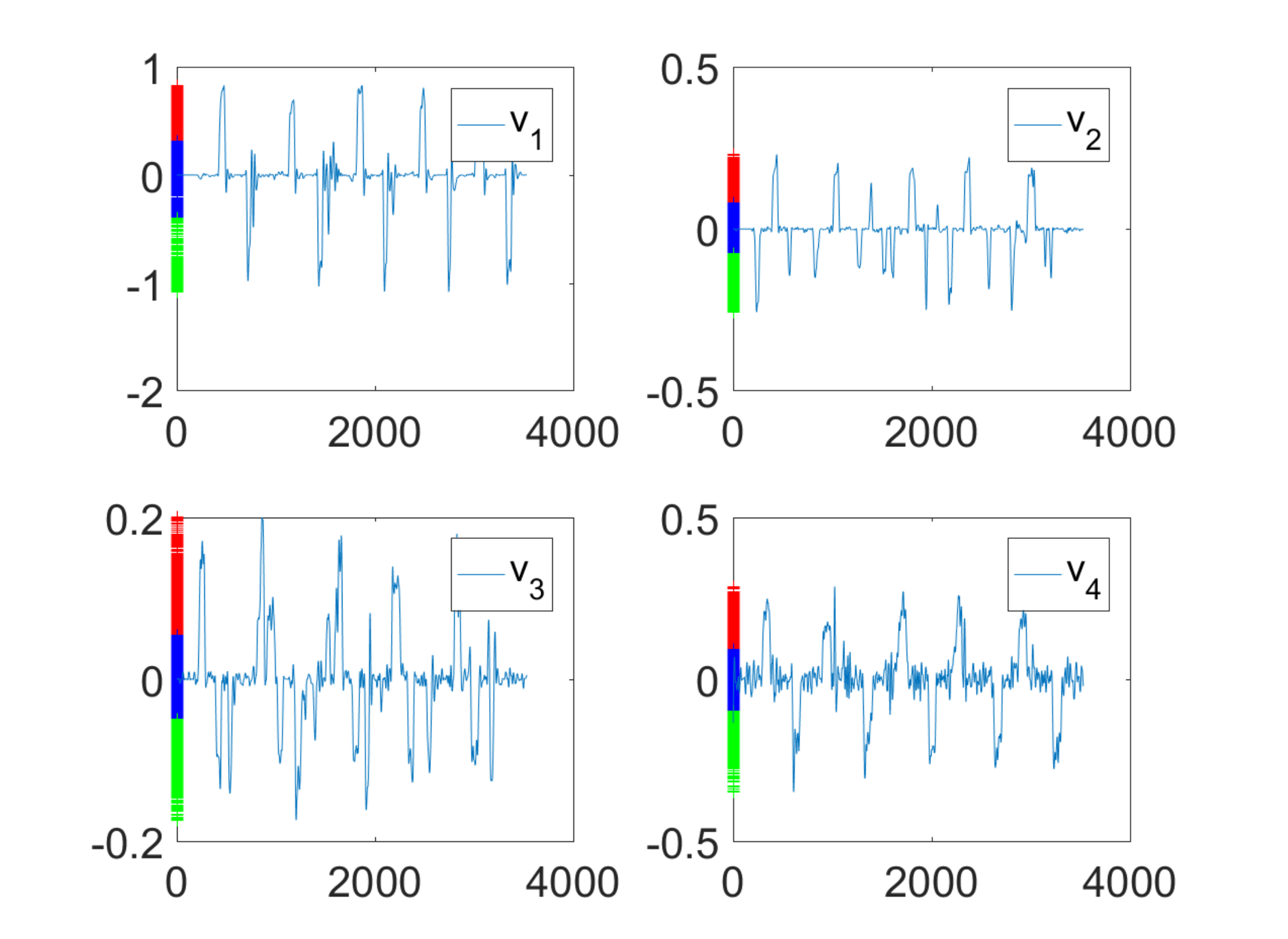}
 	\caption{k-means clustering of sampled velocities for each of the four joints of model Excavator with $ k=3 $. Clustering demarcates the intended three different states for $ \actelem_j $.}
 	\vspace{-0.2in}
 	\label{fig:kmeans}
 \end{figure}

 Let  $ v_\tindex \in \mathbb{R}^{\Njoints \times 1}$ denote the sampled velocity at time $ \tindex $,  $ v_{\tindex j} $ be the velocity of joint $ j $, and $ v_{\cdot j} $ be the set of all sampled velocities $ \{v_{1j},v_{2j},\dots,v_{Tj} \} $ of the joint $ j $. We cluster velocities in the set $  v_{\cdot j} 
 $ using k-means algorithm by setting $ k=3 $ to obtain three clusters corresponding to $ \actelem_j \in \{1,2,3\} $. Our results show that applying k-means consistently produces clusters such as one shown in figure \ref{fig:kmeans}, which clearly demarcates the intended three different states for $ \actelem_j $. From cluster members we calculate the mean $ \mu_{ji} $ and variance $ \sigma_{ji}^2 $ for each cluster $ i\in \{1,2,3\} $. Based on these cluster parameters we define  the probability
 \begin{equation}
 p(\actelem_j = i|v_{\tindex j} ) = \mathcal{N}(v_{\tindex j}|\mu_{ji},\sigma_{ji}^2)
 \end{equation}
 and assign $ \actelem_j $ as follows
 \begin{eqnarray}\label{eq:class}
 \actelem_j=\begin{cases}
 2 \qquad & p(\actelem_j = 2) > \eta\\
 1 \qquad & p(\actelem_j = 1) > p(\actelem_j = 3)\\
 3 \qquad & p(\actelem_j = 3) < p(\actelem_j = 1) \\
 \end{cases}
 \end{eqnarray} 
 where $ \eta $ is a threshold which can be set using a labeled trajectory data. This procedure is repeated for each joint $ j $ to obtain the action primitive $ \actprim_\tindex $ at each time instant $ \tindex $. This process generates action primitive segments $ \mathcal{A}_i $ comprised of similar action primitive $ a_\tindex = a_i $ observed continuously over time instants $ \tindex = \{\tindex_{i},\tindex_{i+1}, \dots,\tindex_j \} $.  An example segmentation (discussed later in details) is shown in figure \ref{fig:segments}, where each colored segment is an action primitive segment ($ \A_i $) that has a particular action primitive ($ a_i $). Total of $ 14 $ unique action primitives (or action primitive segments) were discovered in a truck loading task performed using a model excavator having four revolute joints. Repetition of these unique action primitive segments generates the entire task.  We associate each action primitive segment $ \A_i $ with the corresponding end-effector pose $ s_i $ at the beginning of an action primitive. Each action primitive segment can then be represented by the pair $ (s_i,a_i) $ where $ a_i $ is the associated action primitive. Thus using segmentation procedure we obtain the set state-action pairs $ \obsset = \{(s_1,a_1),(s_2,a_2),\dots, (s_N,a_N) \} $, where $ N $ is the total number of action primitive segments observed in a demonstration. The set $ \obsset $ is used as an input to the inverse reinforcement learning described in the next section. 
 \vspace{-0.2in}
 \vspace{-0.2in}
 \subsection{Dirichlet Process Means Inverse Reinforcement Learning (DPMIRL)}\label{sec:dpmirl}
 We had noted earlier that humans are good at decomposing a given ill-defined task into a series of actionable subgoals. Such decomposition is
 implicit within the human mind, and humans are not always able to clearly explain how they arrived at the
 decomposition \cite{nokes:2010:problem}, \cite{harre2011development}. However, the key idea that is leveraged is to decompose complex task space into subgoals. This is achieved through DPMIRL formulation: a reward based partitioning of task space. Additionally action primitive segments are associated with the inferred subgoals, such that their proper sequencing generates instructions to translate from one subgoal to another. 
 
 We first propose a simplification of action likelihood approximation used in the BNIRL approach (discussed in section \ref{sec:bnirl}). This simplification allows partitioning of state-action pairs based on euclidean distance metric using Dirichlet process means  \cite{kulis2012revisiting}. Typically in an IRL problem entire state space is modeled as an MDP, whereas we define an $ MDP\backslash R$ as a tuple $ \langle S,A,T,\gamma,D\rangle $ where the sets $ S $ and $ A $ consists of the state action pairs in the set $ \obsset $ which is obtained from action primitive based segmentation. This step significantly reduces the computational burden, and potentially renders the analysis of any problem in high dimensional continuous state action space feasible. We simplify the likelihood term in equation (\ref{eq:act_comp}), where the likelihood of an action $ a_i $ w.r.t to a subgoal $ g_{\assignInd_i} $ is computed as $ P(a_i|s_i,\assignInd_i) \propto e^{\alpha ||a_i - a_{\scriptsize CL}||_2} $. As proposed in \cite{michini2013scalable},  $ a_{CL} $ is the action of a closed-loop controller that attempts to go from $ s_i $ to subgoal $ g_{\assignInd_i} $. A simple controller that would generate an action $ a_{CL} $ to reduce the pose error between the subgoal $ g_{\assignInd_i} $  and the present state $ s_i $ is $ a_{CL} \propto (g_{\assignInd_i} - s_i) $. Since a demonstrator is invariably reducing the pose error between his/her state $ s_i $ and the subgoals, we argue that the action $ a_i $ in the demonstration is also proportional to the pose error. Hence given a particular subgoal $ g_{\assignInd_i} $, even $ a_i \propto (g_{\assignInd_i} - s_i)$. Substituting in equation (\ref{eq:act_comp}) we obtain
 \begin{eqnarray}
 P(a_i|s_i,\assignInd_i) & \propto  &e^{\alpha ||\kappa (g_{\assignInd_i} - s_i) - \lambda(g_{\assignInd_i} - s_i)||_2} \\
 & \propto & e^{\alpha' ||(g_{\assignInd_i} - s_i)||_2}
 \end{eqnarray} 
 where we let $ \kappa $ and $ \lambda $ to be scalar proportionality constants, and thus the original action comparison reduces to pose error comparison between the current state $ s_i $ and the subgoals. This result is very intuitive in the sense that any state-action pair $ (s_i, a_i) $ will be partitioned or assigned to the closest subgoal. Thus any action $ a_i $ from a state $ s_i $, is likely to either attain the assigned subgoal or recede away towards the next. This notion is utilized to partition the state action pairs in the set $ \obsset $, using euclidean distance metric on state locations i.e. $ ||s_i||_2 $ for a state $ s_i $. This is performed using DP-means algorithm discussed in section \ref{sec:dpmeans}. Thus we  obtain clusters $ \{c_1,\dots,c_k\} $ where the number of clusters $ k $ is not known a-priori, and each cluster $ c_j $ consists of member states $ \{s_i \in c_j:\assignInd_i = j \} $. Since each state $ s_i \in c_j$ is associated with an action primitive $ a_i $, we define a set $ \mathcal{S}_j  $ w.r.t the cluster $ c_j $ as $  \{  (s_i,a_i)\in \mathcal{S}_j:s_i \in c_j \} $. We define the subgoal as a multivariate Gaussian $ X_j \sim \mathcal{N}(\mu_{j},\Sigma_j) $ in $ n $-dimensional space (formed by end-effector position and mechanism) whose parameters are obtained from the member states of cluster $ c_j $. 
 
 To ensure that the instruction policy for a given task, is generalizable to a novel task configuration it is important to define appropriate reference coordinate frames to compute the euclidean norm. Given a task configuration we define a coordinate frame centered on each known object. Let $ M$ be the number of objects and $ m_j $ be the center of the $ j^{th} $ object's coordinate frame w.r.t the base frame of the robot. We divide the states in the set $ S $ into $ M $ disjoint sets $ \{S_1,\dots,S_M\} $, where each $ s_i $ is assigned to $ S_j $ such that $ \arg \min_j ||s_i - m_j||_2  $. We run DP-means separately for each set $ S_j $, to obtain subgoals as described in algorithm \ref{alg:dpmirl}.  Thus we have decomposed the task space into finite subgoals, most importantly defining subgoals as a multivariate Gaussian allows us to evaluate the likelihood of any state $ s $ in space w.r.t to each subgoal, to determine the most likely subgoal. This assignment allow us to select sequence of action primitives as instructions to reach next subgoal, this is discussed next.
 \begin{algorithm}[tbh]
 	\caption{DPMIRL}
 	\label{alg:dpmirl}
 	\begin{algorithmic}
 		\STATE {\bfseries Input:} Action primitive segments $ (S,A) \in \obsset $, $ M $ objects of interest centered at $ m_j $.
 		\STATE - Assign state $ s_i $ to $ S_j $ s.t. $ \arg \min_j ||s_i - m_j||_2 $ 
 		\STATE - Run DP-means algorithm for states in each $ S_j $, to obtain set of clusters $ C_j =\{c_{j1},\dots,c_{jk} \}$ 
 		\STATE - Set of subgoals $ X = \emptyset $, $ p = |X| = 0 $
 		\FOR {each set of clusters $C_j$} 
 		\FOR {each cluster $ k $ in $ C_j $}
 		\STATE - Compute mean $ \mu_{kj} $ and variance $ \Sigma_{kj}  $ for the member states $ s \in c_{jk} $ 
 		\STATE - Add $ X_p \sim \mathcal{N}(\mu_{kj},\Sigma_{kj}) $ to set $ X $, increment $ p $,
 		\ENDFOR
 		\ENDFOR
 		\STATE {\bfseries Output:} Set of subgoals $ X $.
 	\end{algorithmic}
 \end{algorithm}
 
 \begin{figure*}[h]
 	\centering
 	\begin{tikzpicture}[
 	node distance=4mm,
 	title/.style={font=\fontsize{12}{12}\color{black}\ttfamily},
 	typetag/.style={rectangle, draw=black!50, font=\scriptsize\ttfamily}
 	]
 	\draw (-2.5,-0.5) rectangle (15,-4);
 	
 	\node (x1) at (-1,-1) {$ X_1 $};
 	\draw [draw=black] (x1) circle (0.3cm);
 	\draw[dashed] (-1,-1.3) -- (-2.2,-2.1) -- (1.3,-2.1) -- cycle;

 	\node (z1) at (-2,-2.5) {$ Z^1_1 $};
 	\draw [draw=black] (z1) circle (0.3cm);
 	\node (decomp) [title] at (6,-0.4) {  };
 	\node (z2) [right=of z1, xshift=0mm, yshift=0mm]{$ Z^1_2 $};
 	\draw [draw=black] (z2) circle (0.3cm);
 	\node (z3) [right=of z2, xshift=0mm, yshift=0mm]{$ \dots $};
 	\node (z4) [right=of z3, xshift=0mm, yshift=0mm]{$ Z^1_{T_1} $};
 	\draw [draw=black] (z4) circle (0.3cm);
 	
 	\draw[->] (z1.east)--(z2.west);
 	\draw[->] (z2.east)--(z3.west);
 	\draw[->] (z3.east)--(z4.west);
 	
 	\draw[-latex] (x1.south)--(z1.north);
 	\draw[-latex] (x1.south)--(z2.north);
 	\draw[-latex] (x1.south)--(z3.north);
 	\draw[-latex] (x1.south)--(z4.north);
 	
 	\node (y1) [below=of z1, xshift=0mm, yshift=0mm] {$ y^1_1 $};
 	\draw [draw=black] (y1) circle (0.29cm);
 	\node (y2) [below=of z2, xshift=0mm, yshift=0mm]{$ y^1_2 $};
 	\draw [draw=black] (y2) circle (0.29cm);
 	\node (y3) [below=of z3, xshift=0mm, yshift=0mm]{$ \dots $};
 	\node (y4) [below=of z4, xshift=0mm, yshift=0mm]{$ y^1_{T_1} $};
 	\draw [draw=black] (y4) circle (0.29cm);
 	
 	\draw[-latex] (z1.south)--(y1.north);
 	\draw[-latex] (z2.south)--(y2.north);
 	\draw[-latex] (z4.south)--(y4.north);

 	\node (x2) at (3.5,-1) {$ X_2 $};
 	\draw [draw=black] (x2) circle (0.3cm);
 	\draw[dashed] (3.5,-1.3) -- (2.2,-2.1) -- (6,-2.1) -- cycle;

 	\node (z1n) at (2.5,-2.5) {$ Z^2_1 $};
 	\draw [draw=black] (z1n) circle (0.3cm);

 	\node (z2) [right=of z1n, xshift=0mm, yshift=0mm]{$ Z^2_2 $};
 	\draw [draw=black] (z2) circle (0.3cm);
 	\node (z3) [right=of z2, xshift=0mm, yshift=0mm]{$ \dots $};
 	\node (z4) [right=of z3, xshift=0mm, yshift=0mm]{$ Z^2_{T_2} $};
 	\draw [draw=black] (z4) circle (0.3cm);
 	
 	\draw[->] (z1n.east)--(z2.west);
 	\draw[->] (z2.east)--(z3.west);
 	\draw[->] (z3.east)--(z4.west);
 	\draw[->] (x1.east)--(x2.west);

 	\draw[-latex] (x2.south)--(z1n.north);
 	\draw[-latex] (x2.south)--(z2.north);
 	\draw[-latex] (x2.south)--(z3.north);
 	\draw[-latex] (x2.south)--(z4.north);
 	
 	\node (y1) [below=of z1n, xshift=0mm, yshift=0mm] {$ y^2_1 $};
 	\draw [draw=black] (y1) circle (0.29cm);
 	\node (y2) [below=of z2, xshift=0mm, yshift=0mm]{$ y^2_2 $};
 	\draw [draw=black] (y2) circle (0.29cm);
 	\node (y3) [below=of z3, xshift=0mm, yshift=0mm]{$ \dots $};
 	\node (y4) [below=of z4, xshift=0mm, yshift=0mm]{$ y^2_{T_2} $};
 	\draw [draw=black] (y4) circle (0.29cm);
 	
 	\draw[-latex] (z1n.south)--(y1.north);
 	\draw[-latex] (z2.south)--(y2.north);
 	\draw[-latex] (z4.south)--(y4.north);

 	\node (x3) at (7,-1) {$ \dots $};
 	\draw[->] (x2.east)--(x3.west);
 	\node(x3t) [below=of x3, xshift=0mm, yshift=-7.5mm]{$ \dots $};
 	
 	\node (x4) at (9,-1) {$ X_f $};
 	\draw [draw=black] (x4) circle (0.3cm);
 	\draw[dashed] (9,-1.3) -- (7.8,-2.1) -- (11.3,-2.1) -- cycle;
 	\draw[->] (x3.east)--(x4.west);
 	
 	\node (sub) [right=of x4, xshift=22mm, yshift=0mm]{\textit{Subgoals} };
 	\node (Act) [right=of x4, xshift=22mm, yshift=-12mm]{ \textit{Action Primitives} };
 	\node (Act1) [below=of Act, xshift=0mm, yshift=-3mm]{ \textit{Acutator Velocities} };

 	\node (z2n) at (8,-2.5) {$ Z^f_1 $};
 	\draw [draw=black] (z2n) circle (0.3cm);
 	\node (z2) [right=of z2n, xshift=0mm, yshift=0mm]{$ Z^f_2 $};
 	\draw [draw=black] (z2) circle (0.3cm);
 	\node (z3) [right=of z2, xshift=0mm, yshift=0mm]{$ \dots $};
 	\node (z4) [right=of z3, xshift=0mm, yshift=0mm]{$ Z^f_{T_f} $};
 	\draw [draw=black] (z4) circle (0.3cm);
 	
 	\draw[->] (z2n.east)--(z2.west);
 	\draw[->] (z2.east)--(z3.west);
 	\draw[->] (z3.east)--(z4.west);

 	\draw[-latex] (x4.south)--(z2n.north);
 	\draw[-latex] (x4.south)--(z2.north);
 	\draw[-latex] (x4.south)--(z3.north);
 	\draw[-latex] (x4.south)--(z4.north);
 	
 	\node (y1) [below=of z2n, xshift=0mm, yshift=0mm] {$ y^f_1 $};
 	\draw [draw=black] (y1) circle (0.29cm);
 	\node (y2) [below=of z2, xshift=0mm, yshift=0mm]{$ y^f_2 $};
 	\draw [draw=black] (y2) circle (0.29cm);
 	\node (y3) [below=of z3, xshift=0mm, yshift=0mm]{$ \dots $};
 	\node (y4) [below=of z4, xshift=0mm, yshift=0mm]{$ y^f_{T_f} $};
 	\draw [draw=black] (y4) circle (0.29cm);
 	
 	\draw[-latex] (z2n.south)--(y1.north);
 	\draw[-latex] (z2.south)--(y2.north);
 	\draw[-latex] (z4.south)--(y4.north);
 	\end{tikzpicture}
 	\caption{Task-come-Instruction model obtained based on subgoals generated using DPMIRL and the action primitive segments.}
 	\label{fig:taskmodel}
 \end{figure*}
 \begin{figure}
 	\begin{subfigure}[tbh]{0.45\columnwidth}
 		\centering
 		\includegraphics[width=\textwidth]{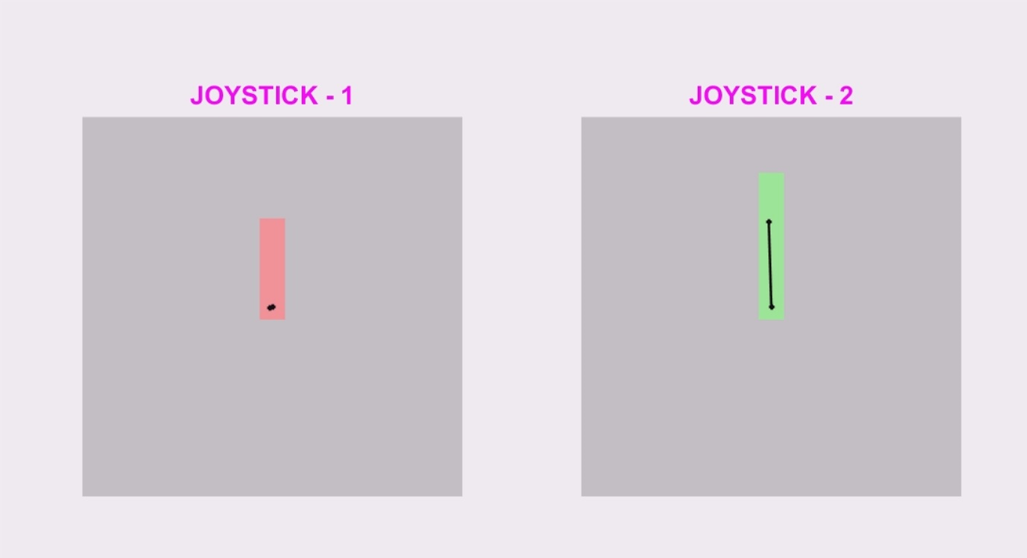}
 		\caption{Interface 1}
 		\label{fig:Barsgui}
 	\end{subfigure}
 	\begin{subfigure}[tbh]{0.45\columnwidth}
 		\centering
 		\includegraphics[width=\textwidth]{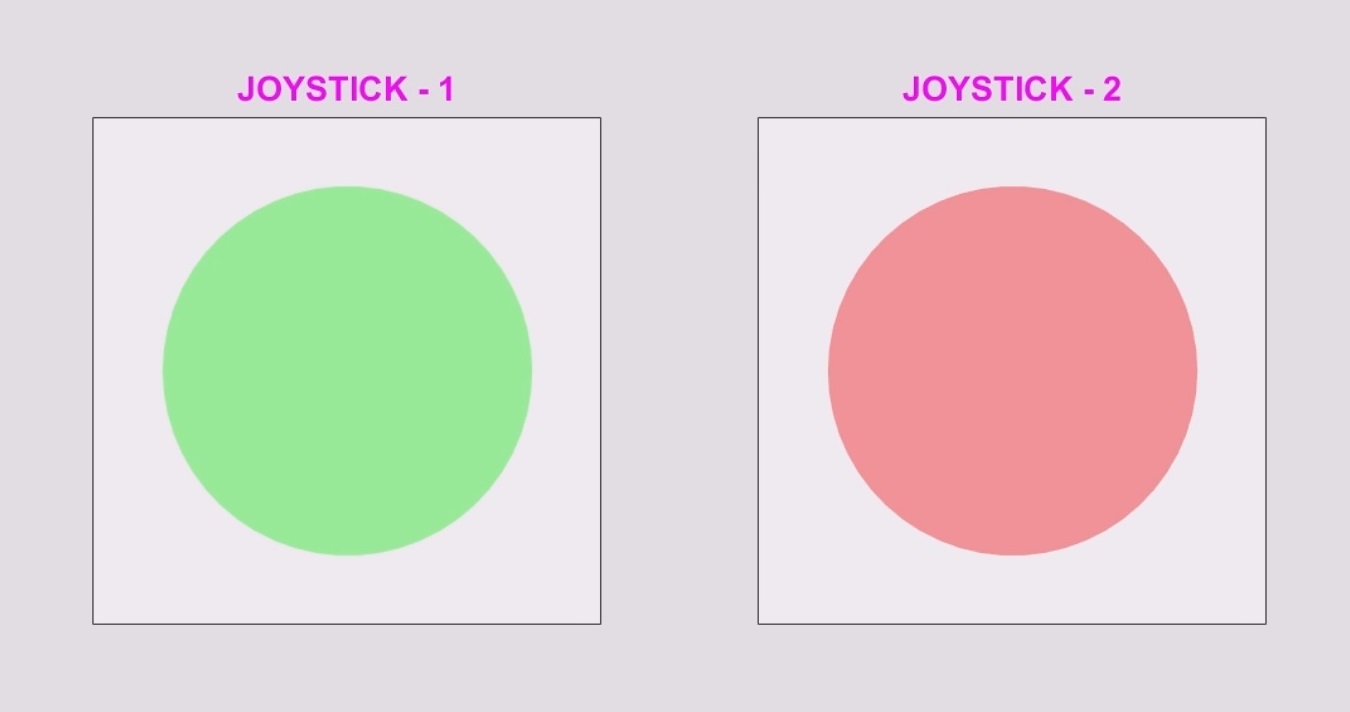}
 		\caption{Interface 2}
 		\label{fig:circlesgui}
 	\end{subfigure}
 	\caption{Visual interface for instructing humans: a) The interface depicts desired joystick movement direction and magnitude using red colored bars with appropriate length. Thin black line corresponds to the actuation by the operator, which when matched, turns the red colored bar to green. b) Two circles correspond to each joystick. They turn green if the operator takes the desired action, and remain red otherwise. The goal is to reinforce desired behavior with minimal communication. 
 	}
 	\label{fig:gui}
 \end{figure}
 \subsection{Task-come-Instruction Model}\label{sec:instructionmodel}
 DPMIRL partitions the task space into subgoals this is analogous to expert human operators who also decompose a given loosely
 defined task (such as leveling a construction site or loading a truck) into a series of actionable subgoals. This analogy is used as a building block for the instruction model that can be used by a robot  to instruct or guide human operator. Action primitive based segmentation plays an important role in such a model, as the transition between these action primitives results in a transfer from one subgoal to another. Further an action primitives can be easily communicated to the human operator as they represent activation of single or multiple actuators which are in turn operated by specific joysticks. An example of action primitive and its corresponding communication in terms of joystick movements via a simple graphical interface is depicted in figure \ref{fig:Barsgui}. 
 
 We now elaborate the model construction which is generated autonomously  based on the subgoals obtained using DPMIRL and the action primitive based segments. Construction of this model has to be such that the task is elaborated in the form of transition among the subgoals. One such construction using hierarchy of Markov chains also known as a Dynamical Bayesian Network is shown in figure \ref{fig:taskmodel}, where the transition between subgoals is modeled by the topmost Markov chain. In this construction, we utilize a fact that each subgoal $ (X_i) $ generated using DPMIRL has an associated set $ \subgStateAct_i $ of state-action primitive pairs, and there exists a Markov chain of variables $ \{Z^{i}_1, \dots,Z^i_{T_i} \} $, where each variable is an action primitive at time instants $ \tindex = 1,\dots,T_i $, that results in a translation to the next subgoal $ X_j $. Hence the Markov chain under each subgoal $ X_i $ models the transition among the action primitives associated with that subgoal. These transition models for action primitives are obtained from the segmentation of demonstration data and counting the transitions between the action primitive segments. The final layer of the model is actuator velocity variable $ y_\tindex^i $ that is conditional on the action primitive $ Z^{i}_\tindex $, and is modeled as a Gaussian distribution over the sampled actuator velocities contained in the action primitive segment. Given the most likely subgoal ($ X_i $) for the current state $ s $ and the previous action primitive $ Z^i_{\tindex-1} $, the generative model for getting instruction in the form of next action primitive $ Z^i_{\tindex} $, using the model in figure \ref{fig:taskmodel}, is as follows:
 \begin{eqnarray}
 P(X_{i+1}|X_{i}) & \sim & \Pi \label{eq:inspol1} \\
 P(Z^i_\tindex | Z^i_{\tindex-1},X_i) & \sim & \pi(X_i) \label{eq:inspol2}\\
 P(y^i_\tindex|Z^i_\tindex) & \sim & F(\theta_{Z^i_\tindex}) \label{eq:inspol3}
 \end{eqnarray}
 and the parameters for this model are the transition distribution $ \Pi $ for subgoals, the subgoal specific transition model $ \pi(X_i) $ for the associated action primitives, and the parameter vector $ \theta_{Z^i_\tindex} $ that models conditional distribution of actuator velocities given the action primitive. 
 
 \subsection{Instruction Interface}\label{sec:interface}
 Human robot interaction (HRI), an entire field of research is devoted to investigate most effective human-robot interfaces. A survey of the techniques typically utilized there is beyond the scope of this work. Rather, in this work we resort to exploratory but exhaustive evaluations using two fundamentally different styles of visual interfaces to communicate the instructional policy learned using methods developed in the previous sections. Our goal is to compare generic reward based interfaces against specific instructional interfaces. 
 Given the nature of uncertainty and human presence around construction equipments, safety and situational awareness of the operator becomes another key aspect for the interface design. It is desirable therefore to have simple visual interface that reinforce desired skills while demanding minimal operator attention. On the other hand, the task is complex, so common wisdom is to err towards providing specific instructions on which actions to take. 
 Accordingly, the first interface indicates the extent of desired actuation, current joystick position along with positive reinforcement by turning red colored bar to green, when an operator takes a desired action. This interface trades off operator's attention to communicate more information. On the other hand, the second interface provides positive reinforcement for desired actuation by changing the color of a circle representing each joystick. It does not provide any other information in terms of the direction on magnitude of joystick motion. 

\begin{figure}[t]
	\centering
	\includegraphics[scale=0.30]{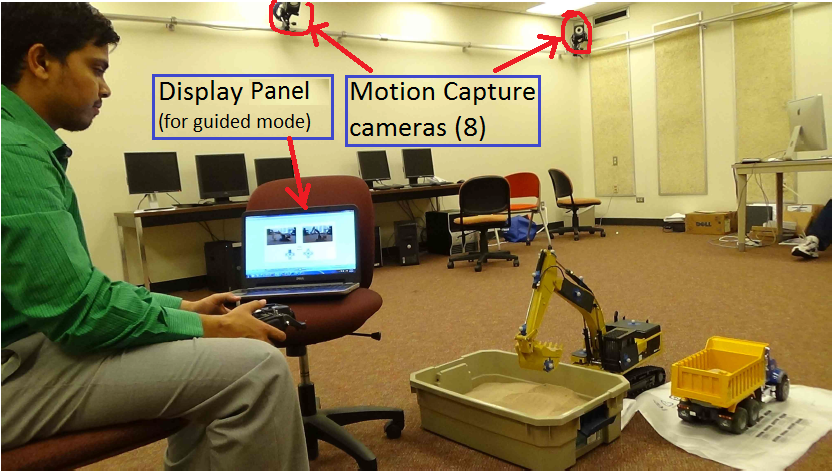}
	\caption{Experimental set-up: Motion capture system, scaled excavator model, and display panel for guided demos.}\label{fig:experiment}
\end{figure}

\begin{figure*}[tbh]
	\centering
	\begin{subfigure}[tbh]{0.32\textwidth}
		\centering
		\includegraphics[width=\textwidth]{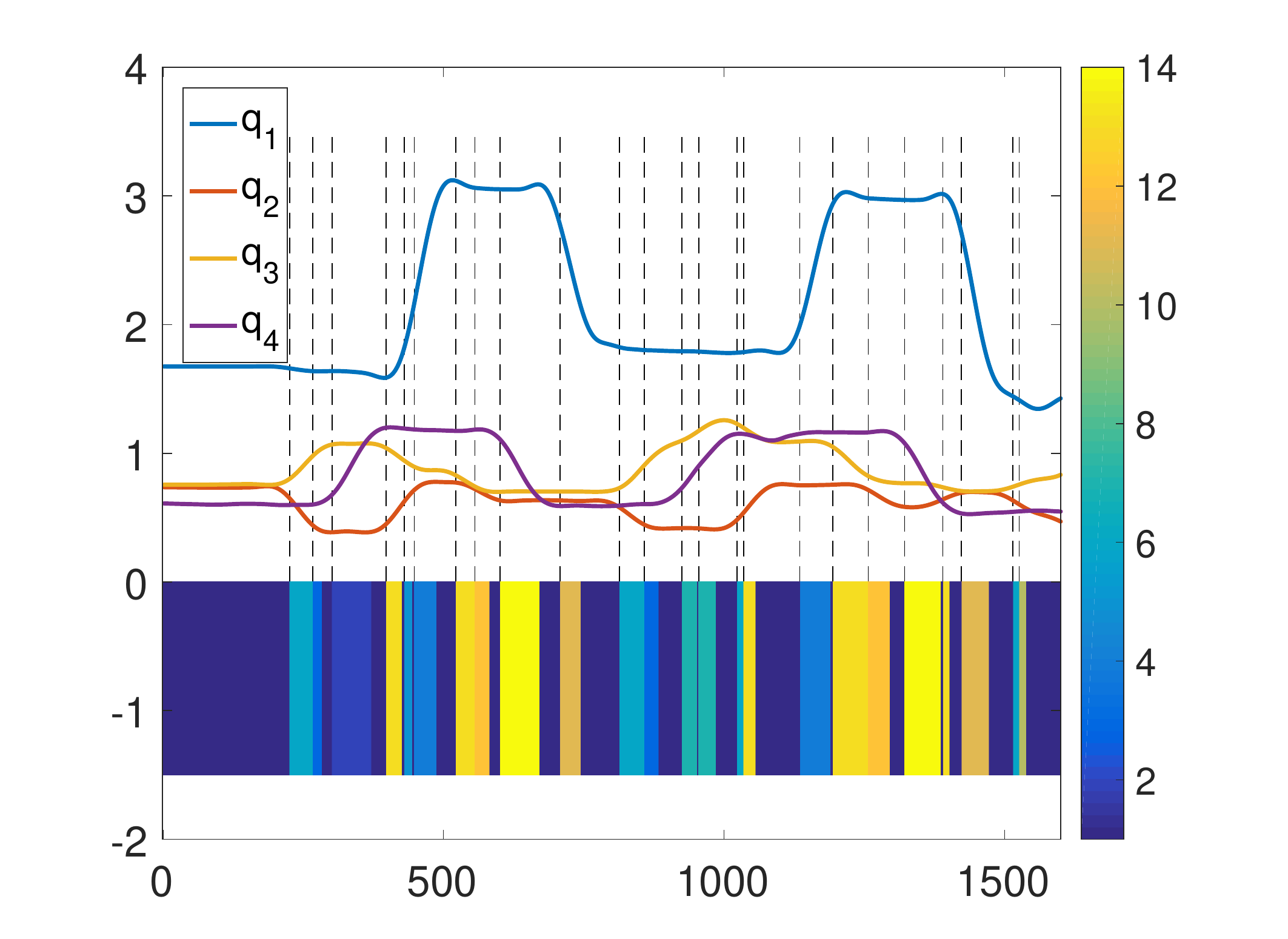}
		\label{fig:seg1}
	\end{subfigure}
	\begin{subfigure}[tbh]{0.32\textwidth}
		\centering
		\includegraphics[width=\textwidth]{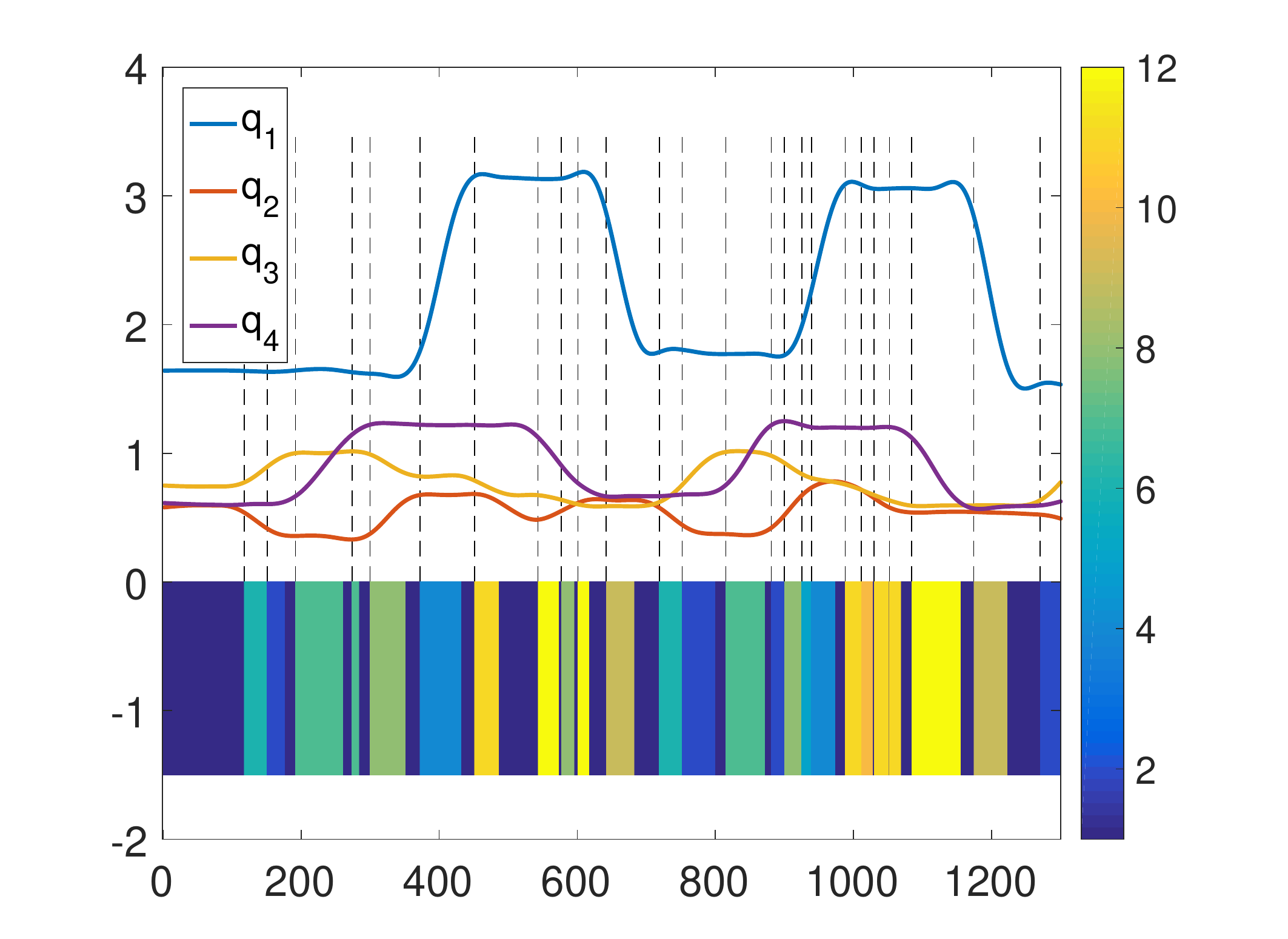}
		\label{fig:seg2}
	\end{subfigure}
	\begin{subfigure}[tbh]{0.32\textwidth}
		\centering
		\includegraphics[width=\textwidth]{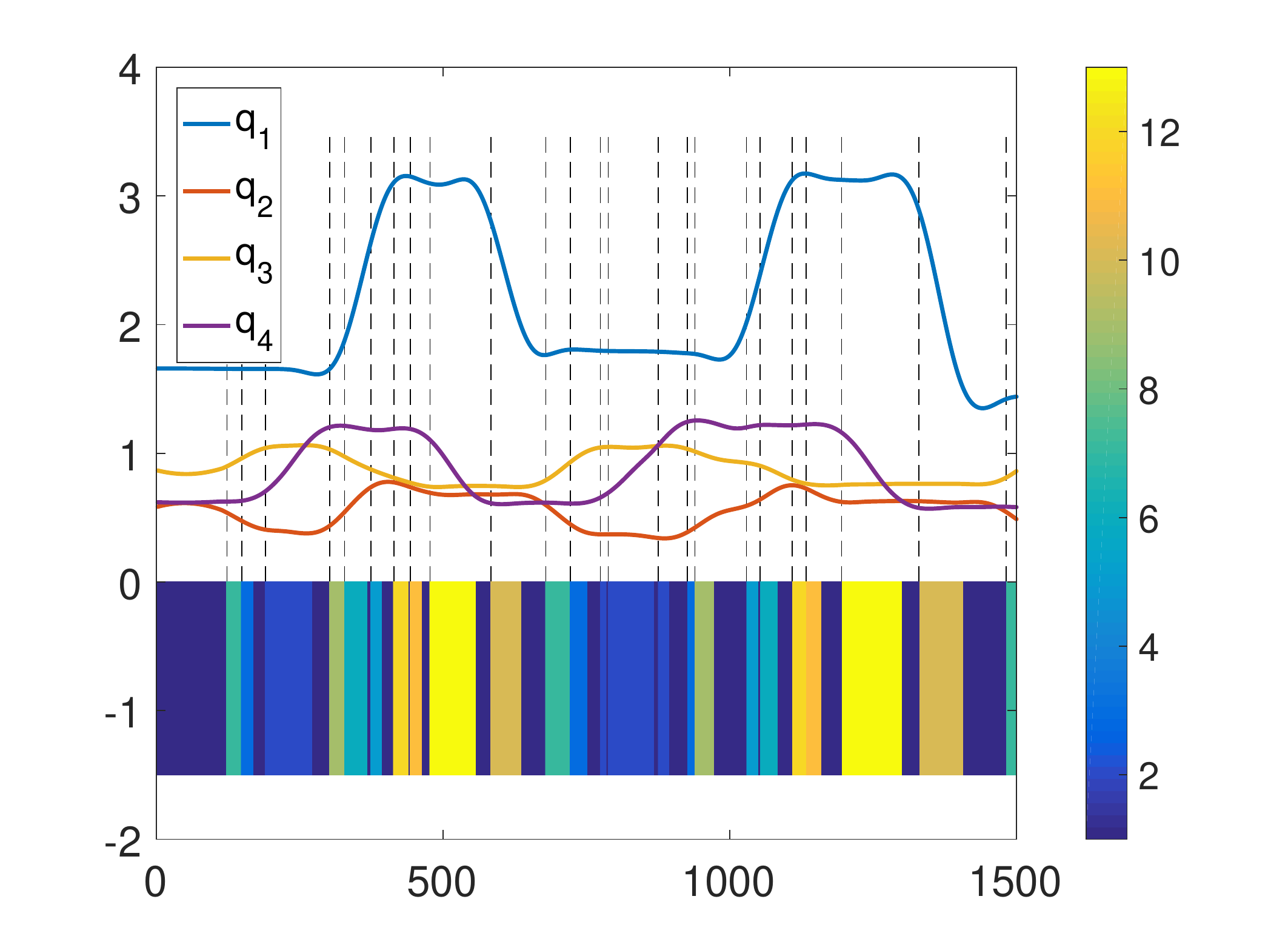}
		\label{fig:seg3}
	\end{subfigure}
	\caption{Segmentations of three demonstration trajectories for two cycles of truck loading task. Each demonstration had five cycles, only two are shown for clarity. Position of each joint $ q_i $ is sampled $ 25 $ times per second 
		Action primitive labels at each time step are indicated by unique colors. Grid lines indicate the starting point of action primitives except for the noisy perturbation action primitive represented in dark blue.}
	\label{fig:segments}
\end{figure*}
\begin{table}[h]
	\caption{Comparison of Action primitive based segmentation with BP-AR-HMM \cite{niekum2013incremental} in terms of computation time}
	\label{table_example}
	\begin{center}
		\begin{tabular}{|c||c|c|c|c|}
			\hline
			& 	 &  &     \\
			No. of Time Series  & 2 & 4 & 6 \\
			\hline
			\hline
			& 	 &  &     \\
			Data size N & 	6690   & 14033 &  23341   \\
			4-D data 	& 	 &  &     \\
			\hline	
			& 	 &  &     \\
			Action Primitive & 21 sec & 41 sec & 68 sec\\
			& 	 &  &    \\
			\hline
			& 	 &  &     \\
			BP-AR-HMM & 1639 sec & 6690 sec & 14752 sec\\
			& 	 &  &    \\
			\hline
		\end{tabular}
	\end{center}
\end{table}
\begin{figure}[tbh]
	\centering
	\begin{subfigure}[tbh]{0.45\columnwidth}
		\centering
		\includegraphics[width=\textwidth]{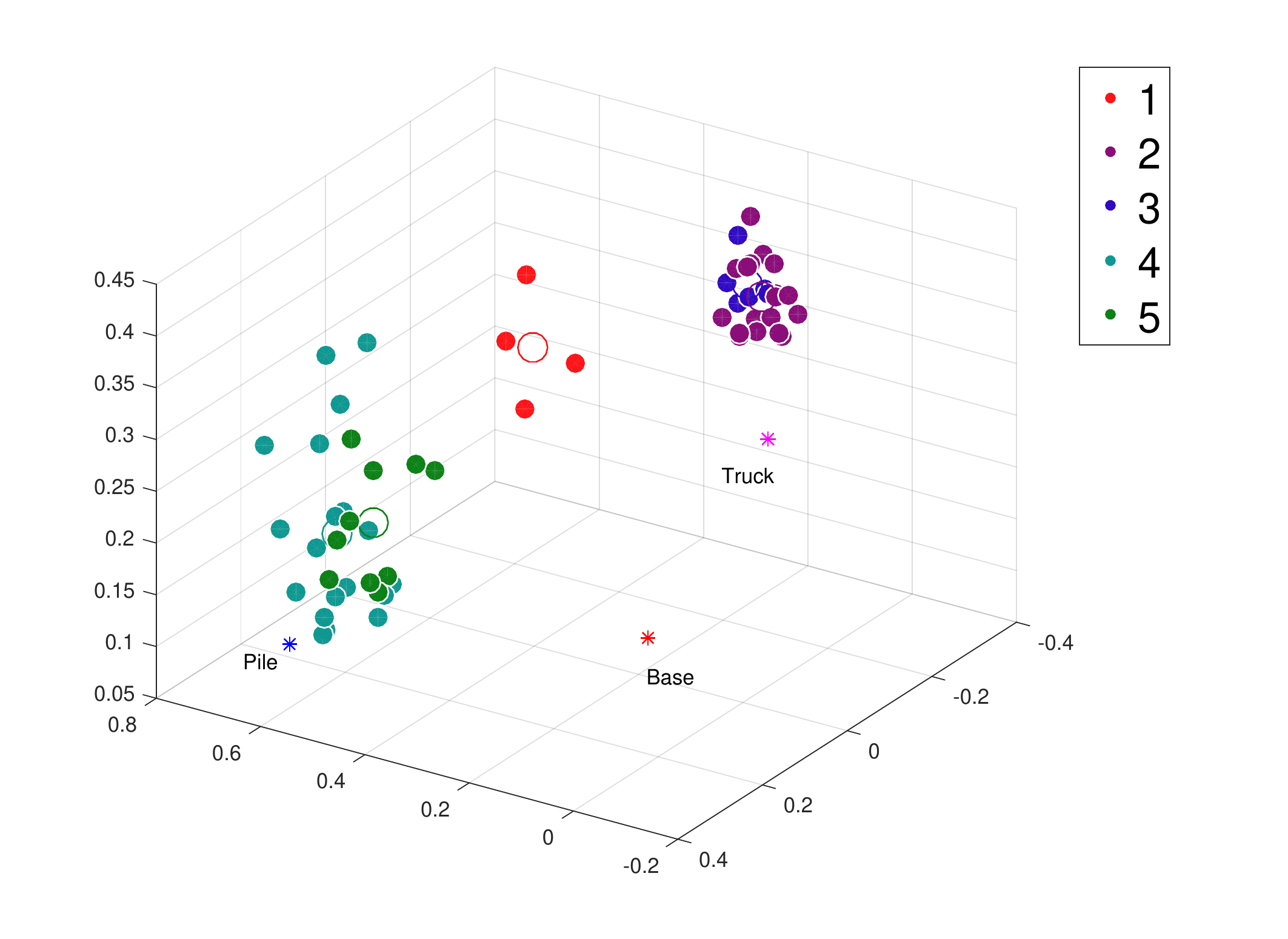}
		\caption{DPMIRL}
		\label{fig:dpmirl}
	\end{subfigure}
	\begin{subfigure}[tbh]{0.45\columnwidth}
		\centering
		\includegraphics[width=\textwidth]{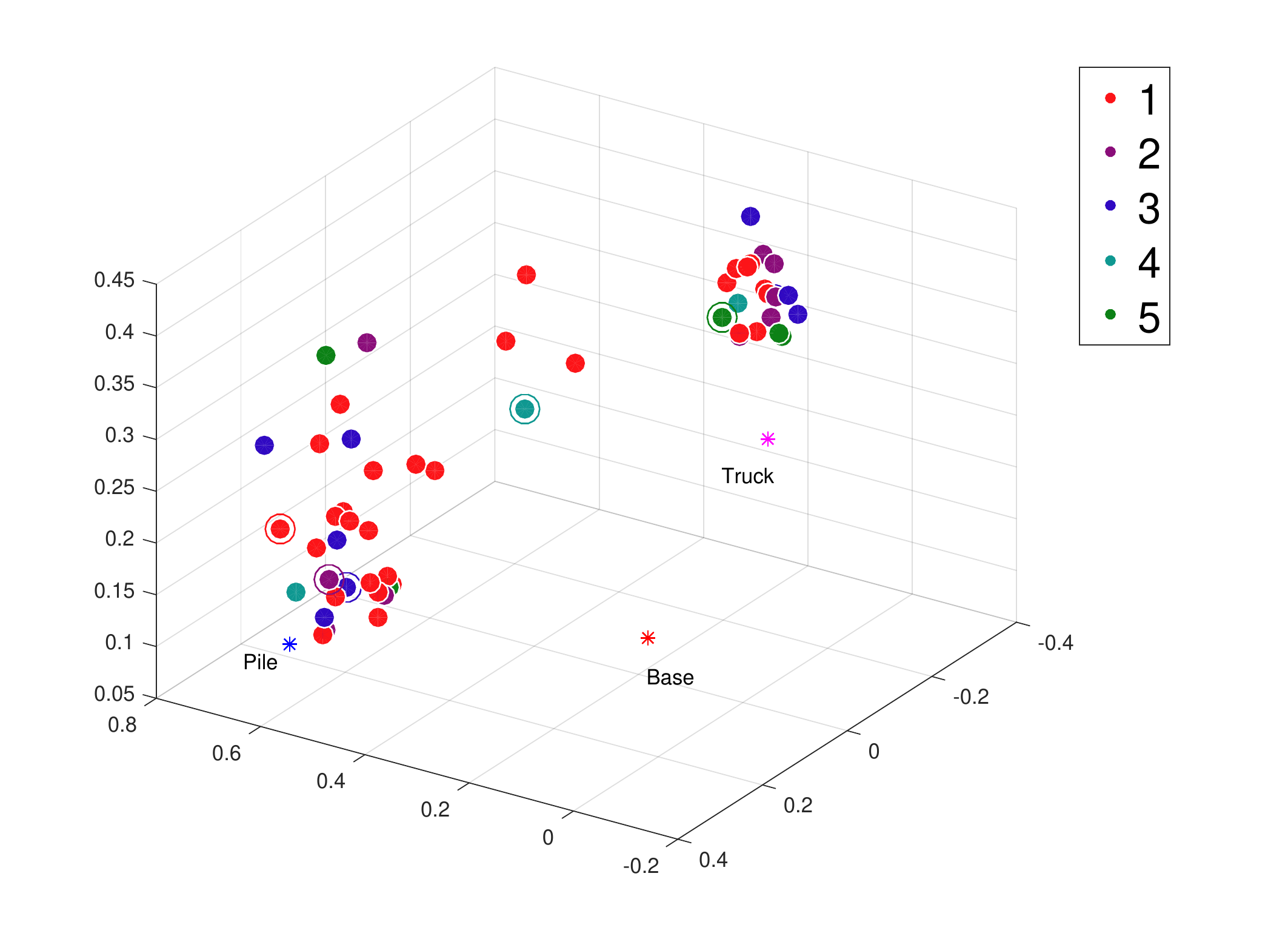}
		\caption{BNIRL}
		\label{fig:bnirl}
	\end{subfigure}
	\caption{Decomposition of task space into subgoals. Task space depicts robot's base, pile and truck location. Both algorithms discovered 5 subgoals shown as big circles in different colors. The action primitive segments associated with each subgoal are highly un-uniform for BNIRL an. Each subgoal is a mean location of its member states shown as filled circles with similar color.}\label{fig:subgoal_oneexp}
\end{figure}
\section{Experiments}
\subsection{Test Platform}
To test the developed approach exhaustive experiments were performed on a $1/14^{th}$ scaled 345D Wedico excavator model a $ 4 $ d.o.f robotic arm manipulator, controlled by a radio transmitter as seen in Figure \ref{fig:experiment}, this model is constructed by Wedico, Germany (http://www.wedico.de/), which specializes in to-scale accurate and fully functional hydraulic construction equipments. Ideally the operator and the display panel shown in the figure \ref{fig:experiment} would be set up inside the excavator but this has been left for future work. However, the results reported here should generalize, furthermore, they are directly applicable to teleoperated robotics. The Wedico excavator lacked joint-angle encoders and internal proprioception, hence all the experiments were performed inside a motion capture facility to ensure real-time data input to the algorithm. In our experiments we record positions and velocities of the four actuators turret, boom, arm and bucket and denote them by vectors $ \posvec $ and $ \vvec $ respectively. We also record end-effector position w.r.t the base frame and bucket angle as four dimensional vector $ \evec $.
\subsection{Learning Instruction Policy}
To demonstrate our approach we selected truck loading task which is a standard
task performed using an excavator, and also happens to be a benchmark operation for fuel consumption analysis (ISO11152) for the excavator family of equipments. Our goal is to learn instruction policy model from demonstrations and evaluate the effectiveness of the teaching interface.  We obtained six set of demonstrations for the truck loading task from a human expert. Each demonstration involved filling up of the truck with sand, for which joint positions $  \posvec  $, joint velocities $ \vvec $ and end-effector position $ \evec $, sampled at $ 25 $Hz were recorded. These demonstration trajectories from the excavator model is used to learn the instruction policy in three steps. 

In the first step we perform action primitive based segmentation of demonstration trajectories, according to the process described in section \ref{sec:segmentation}. Segmentation of three demonstration set is shown in figure \ref{fig:segments}. Each of the figure depict two truck loading cycles for clarity although each demonstration consisted of $ 5-6 $ cycles required to fill the truck to its capacity. We compared our segmentation approach with that of Beta process auto-regressive HMM of \cite{niekum2013incremental} in terms of computation time (table \ref{table_example}) required to segment on a i7-6700K CPU @4GHz and 24 GB RAM machine. Computation time required by BP-AR-HMM scaled geometrically with data size, in comparison action primitive based segmentation scales almost linearly. Being computationally efficient action primitives  can be used to analyze and infer tons of trajectory data from demonstrations. Each action primitive segment $ \A_i $ is a state-action pair $ (s_i,a_i) $ where $ s_i $ is the end-effector pose $ \evec $ from which the action primitive segment began, and $ a_i $ is the associated action primitive. Thus using segmentation procedure we obtain a set of state-action pairs $ \obsset_j = \{(s_1,a_1),(s_2,a_2),\dots, (s_{N_j},a_{N_j}) \} $, where $ N_j $ is the total number of action primitive segments observed in $ j^{th} $ demonstration. Each set $ \obsset_j $ is used as an input to the DPMIRL algorithm, this is the second step which decomposes the task space into finite subgoals. This decomposition is crucial for instruction policy model which consists of action primitive sequence that guides the human operator from one subgoal to another.  One such decomposition that results into six subgoals, three each w.r.t the two task objects (pile and truck) is shown in figure \ref{fig:dpmirl}. Note that clustering is based on euclidean distance metric defined on a four dimensional vector ($ (x,y,z) $ coordinate w.r.t the task object and the bucket angle), which explains the co-location of clusters in the figure, which are actually far apart in the fourth dimension of bucket angle. We followed similar procedure using BNIRL approach \cite{michini2015bayesian} for comparison, and the result is shown in figure \ref{fig:bnirl}. Since the state $ \evec_i $ of action primitive segments  are not co-located with the subgoal location, decomposition obtained using the latter method is not suited for the instruction policy model. Hence as a third step we utilized the decomposition generated by DPMIRL to learn the parameters of instruction policy model (section \ref{sec:instructionmodel}) which is then used to guide human operators in performing the truck loading task.

\subsection{Testing Instruction Policy Model}\label{sec:testing}
We tested the efficacy of the proposed instruction policy model by guiding novice operators in the performance of truck loading task in novel configurations. The process of instructing a human operator while performing a task is depicted (in green) in figure \ref{fig:flowchart}. At any given time instant $ \tindex $, task space configuration comprising of end-effector position, dirt pile and truck position, joint positions and base frame position is available to the co-robot. Co-robot then generates the action primitive $  Z^i_{\tindex} $ using the instructional policy model, based on the previous action primitive $  Z^i_{\tindex-1} $ (calculated from joint position history) and the most likely current subgoal $ X_i $ (subgoal closest to the current end-effector position), using equations (\ref{eq:inspol1})-(\ref{eq:inspol3}).  Figure \ref{fig:gui} shows the two different instruction strategies that are used by the robot to communicate the action primitive to the human operator, who then controls the actuator movement through joystick input.

\begin{figure*}[tbh]
	\centering
	\begin{subfigure}[tbh]{0.23\textwidth}
		\centering
		\includegraphics[width=\textwidth]{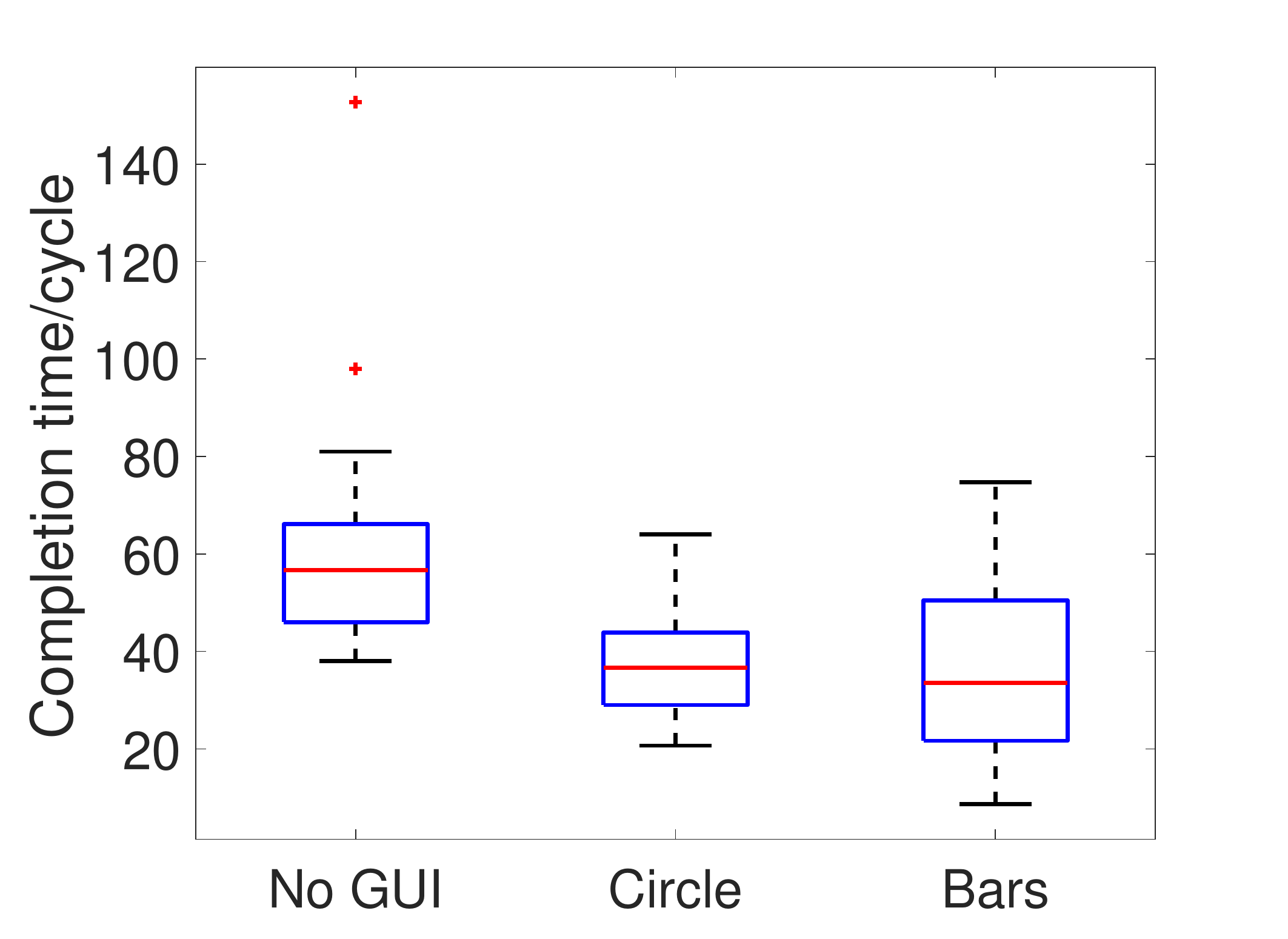}
		\caption{Average Cycle Time}
		\label{fig:b_time}
	\end{subfigure}
	\begin{subfigure}[tbh]{0.23\textwidth}
		\centering
		\includegraphics[width=\textwidth]{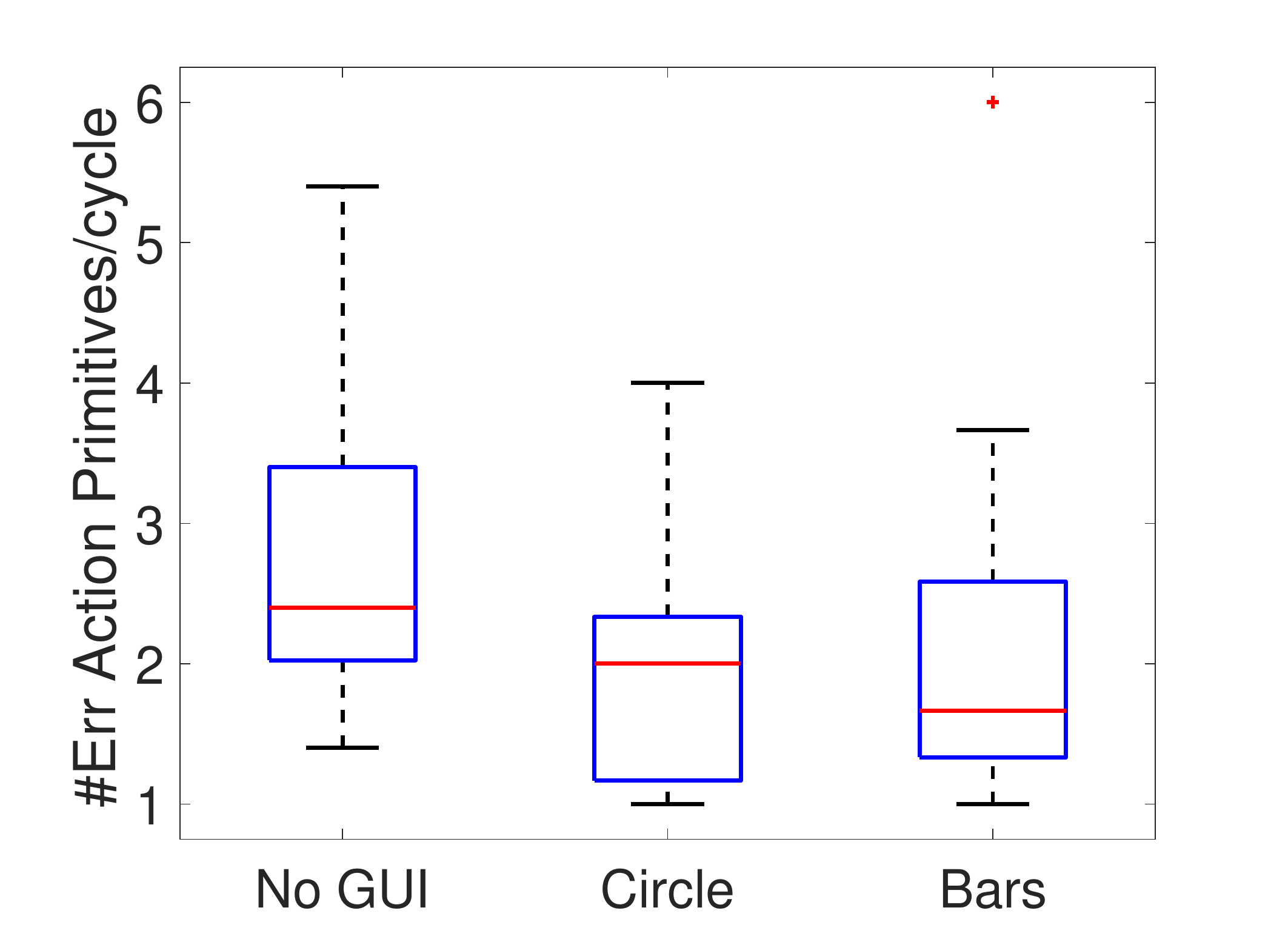}
		\caption{Total Actions}
		\label{fig:b_errors}
	\end{subfigure}
	\begin{subfigure}[tbh]{0.23\textwidth}
		\centering
		\includegraphics[width=\textwidth]{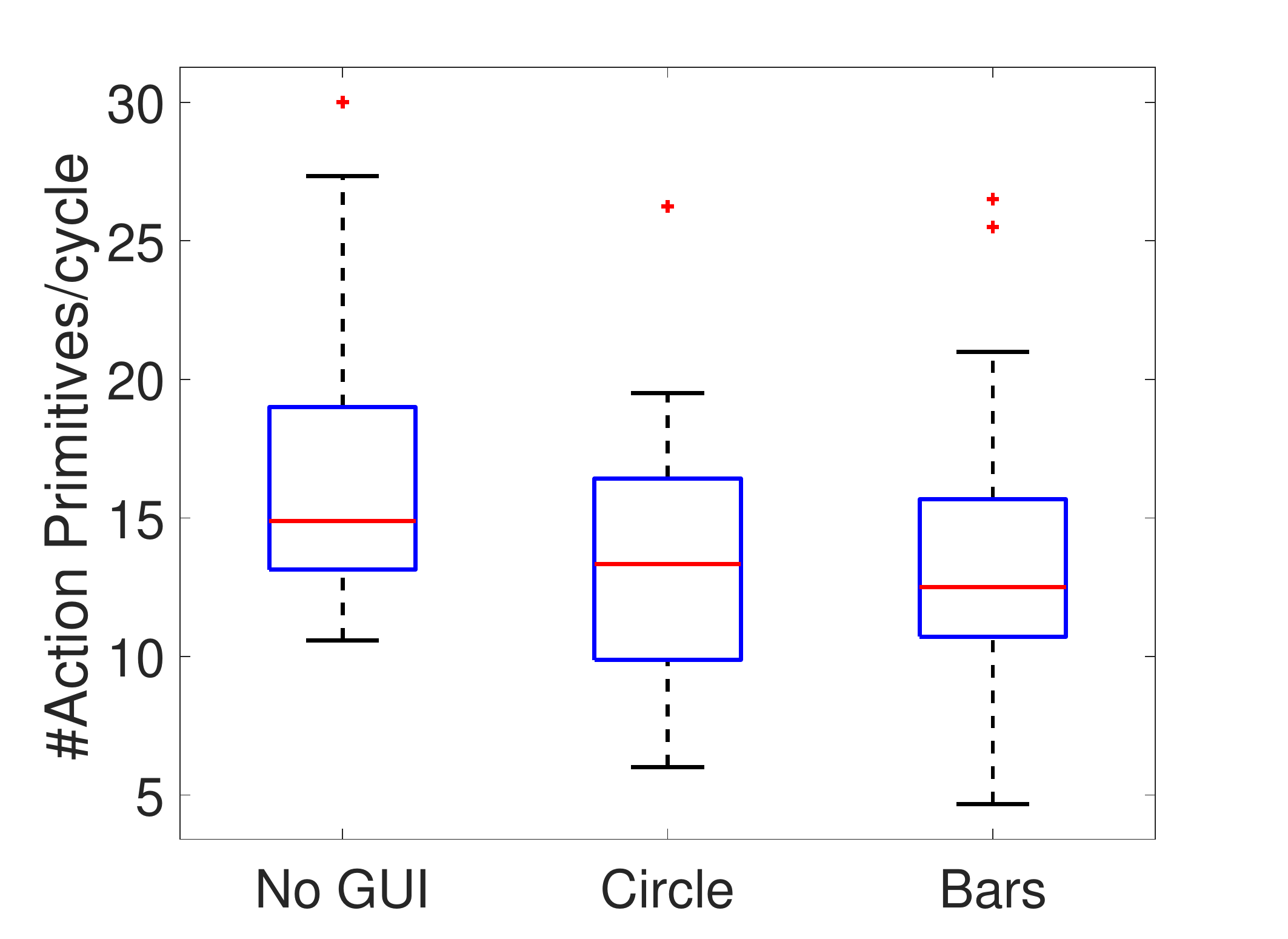}
		\caption{Erroneous Actions}
		\label{fig:b_actions}
	\end{subfigure}
	\begin{subfigure}[tbh]{0.23\textwidth}
		\centering
		\includegraphics[width=\textwidth]{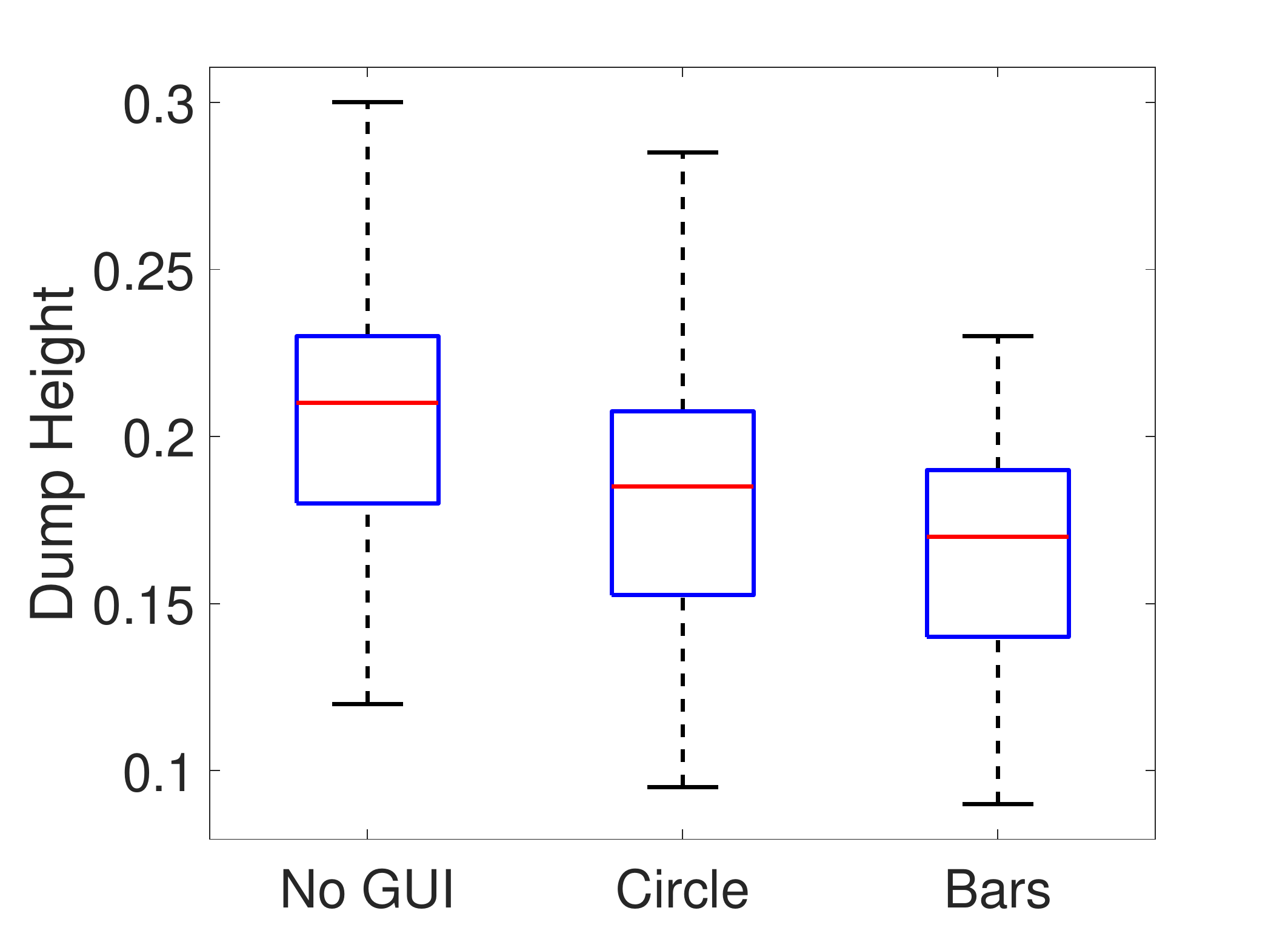}
		\caption{Dump Height}
		\label{fig:b_mhat}
	\end{subfigure}
	\caption{Instruction policy performance results in terms of the four parameters obtained from testing of $ 113 $ participants who randomly distributed into three control groups: i) No GUI, ii) Circle and iii) Bars.}
	\label{fig:boxplots}
\end{figure*}

A total of $ 113 $ participants volunteered under IRB guidelines for the experiments and were split into three groups in order to test the hypothesized visual interfaces. Participants were randomly sorted into groups: Group 1 using the GUI Circles (\ref{fig:circlesgui}), Group 2 using the GUI with Speed Bars (figure \ref{fig:Barsgui}), and Group 3 with no GUI (control).  Each participant was instructed briefly on how the controllers work then given a minimum of three trials to attempt to scoop sand from the tub and deposit into the truck using the controls and the assistance of the selected training GUI.  Participants were timed and videoed and their performance was evaluated in terms of cycle time (completion time for each cycle), number of action primitives executed per cycle, number of erroneous action primitives executed per cycle and the dump height. Dump height is the height above the truck at which the sand is dumped, and is a measure of preciseness. Since each participant performed more than three cycles of truck loading task, an average of these quantities is presented in figure \ref{fig:boxplots}. These results indicate that the instruction policy statistically significantly improves the performance of novice operators across all the parameters. Using instructions from visual interfaces also helped operators in maintaining a lower dump height which is essential in reducing spillage. Interestingly, these results show no significant difference between the two GUI types utilized, a  substantial inference that heavily supports design of simple GUIs that focus on generic rewards rather than complex GUIs that focus on specific instructions. 

\section{Conclusion} 

This paper presented a new method for tackling a class of inverse learning problems in co-robotics where the robot must learn from expert demonstration for teaching non-expert humans. The main contribution of the work presented is a generalizable and scalable technique, that enables the robot to directly learn a instructional policy from expert demonstrations. The paper introduced the notion of action primitives for decomposition and representation of multi-input-multi-output trajectories of complex multi degree of freedom robots. The main advantage of using action primitives, instead of motion primitives which are typically parameterized in the trajectory space, is that action primitives are simpler for the robot to explain to the humans. Furthermore, action primitives can be used as building blocks for a variety of similar tasks utilizing the same base actions and can be generalized to robots with different scales but same joint space, such as other excavators with different boom, bucket, and arm lengths. Finally, action primitives lead to efficient unsupervised clustering of possible robot actions. We demonstrated that utilizing action primitives in a nonparametric unsupervised clustering framework leads to an instructional policy that can utilize current robot pose and subgoal as a feedback signal to provide the correct instruction. Two different interfaces for providing instructional feedback to the human learner were validated in a meticulously constructed large human-robot experiment with 113 human participants. Our results clearly show that there is statistically significant difference in learning rate, task performance times, and retention between guided and un-guided operators. Interestingly, our results also show that with the feedback based instructional policy in place, even simpler operator instruction interfaces perform as well as complex interfaces.

  	
  	\bibliographystyle{plain}
  	\bibliography{daslab_all}
  \end{document}